\documentclass[11pt,a4paper]{article}
\usepackage[hyperref]{emnlp-ijcnlp-2019}
\usepackage{times}
\usepackage{latexsym}
\usepackage{url}

\usepackage{amssymb} 
\usepackage{amsmath}
\usepackage{graphicx} 
\usepackage{multirow}
\aclfinalcopy 

\title{Knowledge-Enriched Transformer for Emotion Detection in Textual Conversations}

\author{Peixiang Zhong$^{1, 2}$, Di Wang$^1$, Chunyan Miao$^{1,2,3}$\\
$^1$Joint NTU-UBC Research Centre of Excellence in Active Living for the Elderly\\
$^2$Alibaba-NTU Singapore Joint Research Institute\\
$^3$School of Computer Science and Engineering\\
Nanyang Technological University, Singapore\\
\texttt{peixiang001@e.ntu.edu.sg}, \texttt{\{wangdi, ascymiao\}@ntu.edu.sg}
}

\date{}

\begin{document}
\maketitle
\begin{abstract}
Messages in human conversations inherently convey emotions. The task of detecting emotions in textual conversations leads to a wide range of applications such as opinion mining in social networks. However, enabling machines to analyze emotions in conversations is challenging, partly because humans often rely on the context and commonsense knowledge to express emotions. In this paper, we address these challenges by proposing a Knowledge-Enriched Transformer (KET), where contextual utterances are interpreted using hierarchical self-attention and external commonsense knowledge is dynamically leveraged using a context-aware affective graph attention mechanism. Experiments on multiple textual conversation datasets demonstrate that both context and commonsense knowledge are consistently beneficial to the emotion detection performance. In addition, the experimental results show that our KET model outperforms the state-of-the-art models on most of the tested datasets in F1 score.
\end{abstract}

\section{Introduction}
\label{introduction}

Emotions are ``generated states in humans that reflect evaluative judgments of the environment, the self and other social agents" \cite{hudlicka2011guidelines}. Messages in human communications inherently convey emotions. With the prevalence of social media platforms such as Facebook Messenger, as well as conversational agents such as Amazon Alexa, there is an emerging need for machines to understand human emotions in natural conversations. This work addresses the task of detecting emotions (e.g., happy, sad, angry, etc.) in textual conversations, where the emotion of an utterance is detected in the conversational context. Being able to effectively detect emotions in conversations leads to a wide range of applications ranging from opinion mining in social media platforms \cite{chatterjee2019understanding} to building emotion-aware conversational agents \cite{zhou2018emotional}. 

\begin{figure}[!t]
\centering
\includegraphics[scale=0.25]{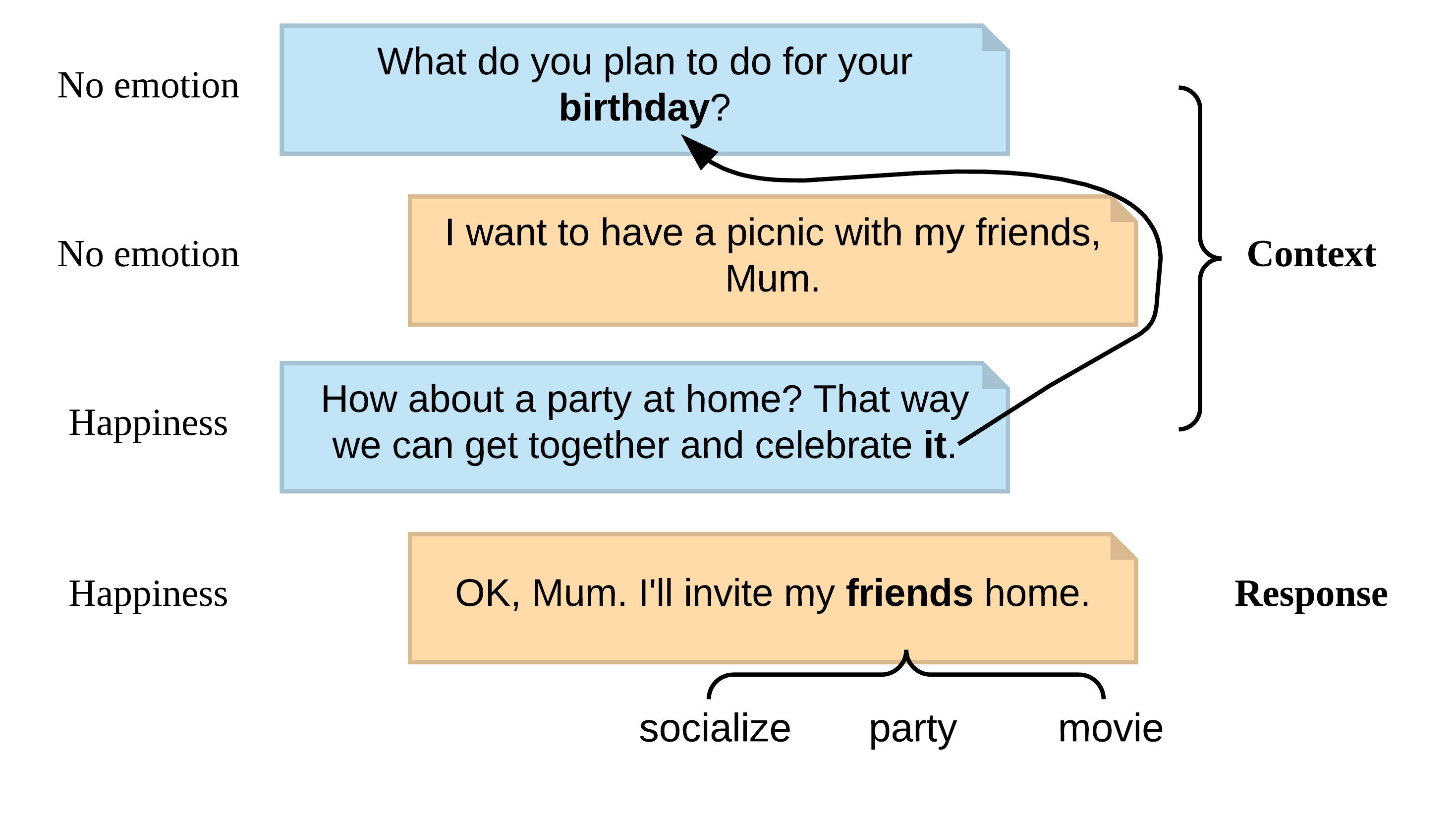}
\caption{An example conversation with annotated labels from the DailyDialog dataset \cite{li2017dailydialog}. By referring to the context, ``it" in the third utterance is linked to ``birthday" in the first utterance. By leveraging an external knowledge base, the meaning of ``friends" in the forth utterance is enriched by associated knowledge entities, namely ``socialize", ``party", and ``movie". Thus, the implicit ``happiness" emotion in the fourth utterance can be inferred more easily via its enriched meaning.}
\label{fig: example}
\end{figure}

However, enabling machines to analyze emotions in human conversations is challenging, partly because humans often rely on the context and commonsense knowledge to express emotions, which is difficult to be captured by machines. Figure \ref{fig: example} shows an example conversation demonstrating the importance of context and commonsense knowledge in understanding conversations and detecting implicit emotions.

There are several recent studies that model contextual information to detect emotions in conversations. \citet{poria2017context} and \citet{majumder2018dialoguernn} leveraged recurrent neural networks (RNN) to model the contextual utterances in sequence, where each utterance is represented by a feature vector extracted by convolutional neural networks (CNN) at an earlier stage. Similarly, \citet{hazarika2018icon, hazarika2018conversational} proposed to use extracted CNN features in memory networks to model contextual utterances. However, these methods require separate feature extraction and tuning, which may not be ideal for real-time applications. 
In addition, to the best of our knowledge, no attempts have been made in the literature to incorporate commonsense knowledge from external knowledge bases to detect emotions in textual conversations. Commonsense knowledge is fundamental to understanding conversations and generating appropriate responses \cite{zhou2018commonsense}.

To this end, we propose a Knowledge-Enriched Transformer (KET) to effectively incorporate contextual information and external knowledge bases to address the aforementioned challenges. The Transformer \cite{vaswani2017attention} has been shown to be a powerful representation learning model in many NLP tasks such as machine translation \cite{vaswani2017attention} and language understanding \cite{devlin2018bert}. The self-attention \cite{cheng2016long} and cross-attention \cite{bahdanau2014neural} modules in the Transformer capture the intra-sentence and inter-sentence correlations, respectively. The shorter path of information flow in these two modules compared to gated RNNs and CNNs allows KET to model contextual information more efficiently. In addition, we propose a hierarchical self-attention mechanism allowing KET to model the hierarchical structure of conversations. Our model separates context and response into the encoder and decoder, respectively, which is different from other Transformer-based models, e.g., BERT \cite{devlin2018bert}, which directly concatenate context and response, and then train language models using only the encoder part.

Moreover, to exploit commonsense knowledge, we leverage external knowledge bases to facilitate the understanding of each word in the utterances by referring to related knowledge entities. The referring process is dynamic and balances between relatedness and affectiveness of the retrieved knowledge entities using a context-aware affective graph attention mechanism. 


In summary, our contributions are as follows:
\begin{itemize}
    \item For the first time, we apply the Transformer to analyze conversations and detect emotions. Our hierarchical self-attention and cross-attention modules allow our model to exploit contextual information more efficiently than existing gated RNNs and CNNs.
    \item We derive dynamic, context-aware, and emotion-related commonsense knowledge from external knowledge bases and emotion lexicons to facilitate the emotion detection in conversations.
    \item We conduct extensive experiments demonstrating that both contextual information and commonsense knowledge are beneficial to the emotion detection performance. In addition, our proposed KET model outperforms the state-of-the-art models on most of the tested datasets across different domains.
\end{itemize}

\section{Related Work}
\label{related work}

\noindent\textbf{Emotion Detection in Conversations:} Early studies on emotion detection in conversations focus on call center dialogs using lexicon-based methods and audio features \cite{lee2005toward, devillers2006real}. \citet{devillers2002annotation} annotated and detected emotions in call center dialogs using unigram topic modelling. 
In recent years, there is an emerging research trend on emotion detection in conversational videos and multi-turn Tweets using deep learning methods \cite{hazarika2018conversational, hazarika2018icon, zahiri2018emotion, chatterjee2019understanding, zhong2019ntuer, poria2019emotion}. \citet{poria2017context} proposed a long short-term memory network (LSTM) \cite{hochreiter1997long} based model to capture contextual information for sentiment analysis in user-generated videos. 
\citet{majumder2018dialoguernn} proposed the DialogueRNN model that uses three gated recurrent units (GRU) \cite{cho2014learning} to model the speaker, the context from the preceding utterances, and the emotions of the preceding utterances, respectively. They achieved the state-of-the-art performance on several conversational video datasets.

\noindent\textbf{Knowledge Base in Conversations:} Recently there is a growing number of studies on incorporating knowledge base in generative conversation systems, such as open-domain dialogue systems \cite{han2015exploiting, asghar2018affective, ghazvininejad2018knowledge, young2018augmenting, parthasarathi2018extending, liu2018knowledge, moghe2018towards, dinan2018wizard, zhong2019affect}, task-oriented dialogue systems \cite{madotto2018mem2seq, wu2018globaltolocal, he2019hierarchical} and question answering systems \cite{kiddon2016globally, hao2017end, sun2018open, mihaylov2018knowledgeable}. 
\citet{zhou2018commonsense} adopted structured knowledge graphs to enrich the interpretation of input sentences and help generate knowledge-aware responses using graph attentions. The graph attention in the knowledge interpreter \cite{zhou2018commonsense} is static and only related to the recognized entity of interest. By contrast, our graph attention mechanism is dynamic and selects context-aware knowledge entities that balances between relatedness and affectiveness.

\noindent\textbf{Emotion Detection in Text:} There is a trend moving from traditional machine learning methods \cite{pang2002thumbs, wang2012baselines, seyeditabari2018emotion} to deep learning methods \cite{abdul2017emonet, zhang2018text} for emotion detection in text. \citet{khanpour2018fine} investigated the emotion detection from health-related posts in online health communities using both deep learning features and lexicon-based features. 

\noindent\textbf{Incorporating Knowledge in Sentiment Analysis:} Traditional lexicon-based methods detect emotions or sentiments from a piece of text based on the emotions or sentiments of words or phrases that compose it \cite{hu2009lyric, taboada2011lexicon, bandhakavi2017lexicon}. 
Few studies investigated the usage of knowledge bases in deep learning methods. \citet{kumar2018knowledge} proposed to use knowledge from WordNet \cite{fellbaum2012wordnet} to enrich the text representations produced by LSTM and obtained improved performance.

\noindent\textbf{Transformer:} The Transformer has been applied to many NLP tasks due to its rich representation and fast computation, e.g., document machine translation \cite{zhang2018improving}, response matching in dialogue system \cite{zhou2018multi}, language modelling \cite{dai2019transformer} and understanding \cite{radford2018improving}. A very recent work \cite{koncel2019text} extends the Transformer to graph inputs and propose a model for graph-to-text generation. 

\begin{figure*}[!t]
\centering
\includegraphics[scale=0.55]{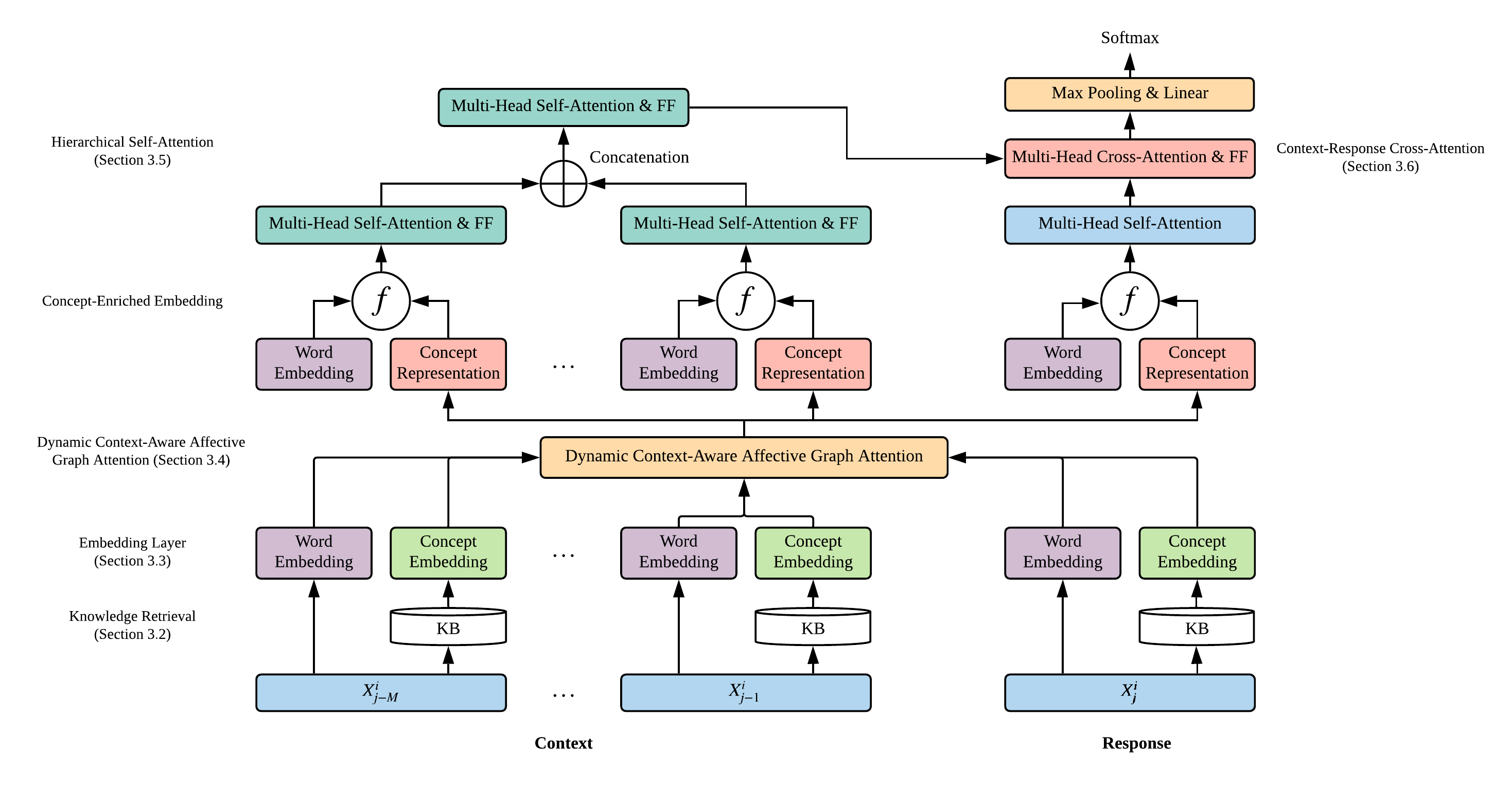}
\caption{Overall architecture of our proposed KET model. The positional encoding, residual connection, and layer normalization are omitted in the illustration for brevity.}
\label{fig: architecture}
\end{figure*}
\section{Our Proposed KET Model}
\label{our model}
In this section we present the task definition and our proposed KET model. 
\subsection{Task Definition}
\label{sec: task definition}
Let $\{X^i_j, Y^i_j\}, i=1,...N, j=1,...N_i$ be a collection of \{\textit{utterance}, \textit{label}\} pairs in a given dialogue dataset, where $N$ denotes the number of conversations and $N_i$ denotes the number of utterances in the $i$th conversation. The objective of the task is to maximize the following function:
\begin{equation}
\label{eqn: objective}
\Phi = \prod_{i=1}^N\prod_{j=1}^{N_i} p(Y^i_j|X^i_j, X^i_{j-1}, ..., X^i_1; \theta),
\end{equation}
where $X^i_{j-1}, ..., X^i_1$ denote contextual utterances and $\theta$ denotes the model parameters we want to optimize. 

We limit the number of contextual utterances to $M$. Discarding early contextual utterances may cause information loss, but this loss is negligible because they only contribute the least amount of information \cite{su2018time}. This phenomenon can be further observed in our model analysis regarding context length (see Section \ref{sec: model analysis}). Similar to \cite{poria2017context}, we clip and pad each utterance $X^i_j$ to a fixed $m$ number of tokens. The overall architecture of our KET model is illustrated in Figure \ref{fig: architecture}.

\subsection{Knowledge Retrieval}
\label{sec: knowledge retrieval}
We use a commonsense knowledge base ConceptNet \cite{speer2017conceptnet} and an emotion lexicon NRC\_VAD \cite{mohammad2018obtaining} as knowledge sources in our model. 

ConceptNet is a large-scale multilingual semantic graph that describes general human knowledge in natural language. The nodes in ConceptNet are concepts and the edges are relations. Each $\langle$\textit{concept1}, \textit{relation}, \textit{concept2}$\rangle$ triplet is an assertion. Each assertion is associated with a confidence score. An example assertion is $\langle$\textit{friends}, \textit{CausesDesire}, \textit{socialize}$\rangle$ with confidence score of 3.46. Usually assertion confidence scores are in the $[1, 10]$ interval. Currently, for English, ConceptNet comprises 5.9M assertions, 3.1M concepts and 38 relations.

NRC\_VAD is a list of English words and their VAD scores, i.e., valence (negative-positive), arousal (calm-excited), and dominance (submissive-dominant) scores in the $[0, 1]$ interval. The VAD measure of emotion is culture-independent and widely adopted in Psychology \cite{mehrabian1996pleasure}. Currently NRC\_VAD comprises around 20K words.

In general, for each non-stopword token $t$ in $X^i_j$, we retrieve a connected knowledge graph $g(t)$ comprising its immediate neighbors from ConceptNet. For each $g(t)$, we remove concepts that are stopwords or not in our vocabulary. We further remove concepts with confidence scores less than 1 to reduce annotation noises. For each concept, we retrieve its VAD values from NRC\_VAD. The final knowledge representation for each token $t$ is a list of tuples: $(c_1, s_1, \textit{VAD}(c_1))$, $(c_2, s_2, \textit{VAD}(c_2))$, ..., $(c_{|g(t)|}, s_{|g(t)|}, \textit{VAD}(c_{|g(t)|}))$, where $c_k \in g(t)$ denotes the $k$th connected concept, $s_k$ denotes the associated confidence score, and $\textit{VAD}(c_k)$ denotes the VAD values of $c_k$. The treatment for tokens that are not associated with any concept and concepts that are not included in NRC\_VAD are discussed in Section \ref{sec: dynamic context-aware affective graph attention}. We leave the treatment on relations as future work.

\subsection{Embedding Layer}
\label{sec: embedding}
 We use a word embedding layer to convert each token $t$ in $X^i$ into a vector representation $\textbf{t} \in \mathbb{R}^{d}$, where $d$ denotes the size of word embedding. To encode positional information, the position encoding \cite{vaswani2017attention} is added as follows:
\begin{equation}
\label{eqn: embed}
\textbf{t} = \textit{Embed}(t) + \textit{Pos}(t).
\end{equation}

Similarly, we use a concept embedding layer to convert each concept $c$ into a vector representation $\textbf{c} \in \mathbb{R}^{d}$ but without position encoding.

\subsection{Dynamic Context-Aware Affective Graph Attention}
\label{sec: dynamic context-aware affective graph attention}
To enrich word embedding with concept representations, we propose a dynamic context-aware affective graph attention mechanism to compute the concept representation for each token. Specifically, the concept representation $\mathbf{c}(t) \in \mathbb{R}^d$ for token $t$ is computed as
\begin{equation}
\label{eqn: concept representation}
\textbf{c}(t) = \sum_{k=1}^{|g(t)|}\alpha_k*\mathbf{c_k},
\end{equation}
where $\mathbf{c_k} \in \mathbb{R}^d$ denotes the concept embedding of $c_k$ and $\alpha_k$ denotes its attention weight. If $|g(t)|=0$, we set \textbf{c}(t) to the average of all concept embeddings. The attention $\alpha_k$ in Equation \ref{eqn: concept representation} is computed as
\begin{equation}
\alpha_k = \textit{softmax}(w_k),
\end{equation}
where $w_k$ denotes the weight of $c_k$.

The derivation of $w_k$ is crucial because it regulates the contribution of $\mathbf{c}_k$ towards enriching $\mathbf{t}$. A standard graph attention mechanism \cite{veličković2018graph} computes $w_k$ by feeding $\mathbf{t}$ and $\mathbf{c}_k$  into a single-layer feedforward neural network. However, not all related concepts are equal in detecting emotions given the conversational context. In our model, we make the assumption that important concepts are those that relate to the conversational context and have strong emotion intensity. To this end, we propose a context-aware affective graph attention mechanism by incorporating two factors when computing $w_k$, namely relatedness and affectiveness. 

\noindent\textbf{Relatedness:} Relatedness measures the strength of the relation between $c_k$ and the conversational context.  The relatedness factor in $w_k$ is computed as 
\begin{equation}
\textit{rel}_k = \textit{min-max}(s_k) * \textit{abs}(\cos(\mathbf{CR}(X^i), \mathbf{c_k})),
\end{equation}
where $s_k$ is the confidence score introduced in Section \ref{sec: knowledge retrieval}, $\textit{min-max}$ denotes min-max scaling for each token $t$, \textit{abs} denotes the absolute function, $\cos$ denotes the cosine similarity function, and $\mathbf{CR}(X^i) \in \mathbb{R}^d$ denotes the context representation of the $i$th conversation $X^i$. Here we compute $\mathbf{CR}(X^i)$ as the average of all sentence representations in $X^i$ as follows:
\begin{equation}
\mathbf{CR}(X^i) = \textit{avg}(\mathbf{SR}(X^i_{j-M}), ..., \mathbf{SR}(X^i_{j})),
\end{equation}
where $\mathbf{SR}(X^i_{j}) \in \mathbb{R}^d$ denotes the sentence representation of $X^i_j$. We compute $\mathbf{SR}(X^i_{j})$ via hierarchical pooling \cite{shen2018baseline} where $n$-gram ($n\leq 3$) representations in $X^i_j$ are first computed by max-pooling and then all $n$-gram representations are averaged. The hierarchical pooling mechanism preserves word order information to certain degree and has demonstrated superior performance than average pooling or max-pooling on sentiment analysis tasks \cite{shen2018baseline}. 

\noindent\textbf{Affectiveness:} Affectiveness measures the emotion intensity of $c_k$. The affectiveness factor in $w_k$ is computed as 
\begin{equation}
\textit{aff}_k = \textit{min-max}(||[\textit{V}(c_k)-1/2, \textit{A}(c_k)/2]||_2),
\end{equation}
where $||.||_k$ denotes $l_k$ norm, $\textit{V}(c_k) \in [0,1]$ and $\textit{A}(c_k) \in [0,1]$ denote the valence and arousal values of $\textit{VAD}(c_k)$, respectively. Intuitively, $\textit{aff}_k$ considers the deviations of valence from neutral and the level of arousal from calm. There is no established method in the literature to compute the emotion intensity based on VAD values, but empirically we found that our method correlates better with an emotion intensity lexicon comprising 6K English words \cite{lrec18mohammad} than other methods such as taking dominance into consideration or taking $l_1$ norm. For concept $c_k$ not in NRC\_VAD, we set $\textit{aff}_k$ to the mid value of $0.5$.

Combining both $\textit{rel}_k$ and $\textit{aff}_k$, we define the weight $w_k$ as follows:
\begin{equation}
\label{eqn: tradeoff}
w_k = \lambda_k * \textit{rel}_k + (1-\lambda_k) * \textit{aff}_k,
\end{equation}
where $\lambda_k$ is a model parameter balancing the impacts of relatedness and affectiveness on computing concept representations. Parameter $\lambda_k$ can be fixed or learned during training. The analysis of $\lambda_k$ is discussed in Section \ref{sec: tradeoff}.

Finally, the concept-enriched word representation $\mathbf{\hat{t}}$ can be obtained via a linear transformation:
\begin{equation}
\mathbf{\hat{t}} = \mathbf{W}[\mathbf{t};\mathbf{c}(t)],
\end{equation}
where $[;]$ denotes concatenation and $\textbf{W} \in \mathbb{R}^{d \times 2d}$ denotes a model parameter. All $m$ tokens in each $X^i_j$ then form a concept-enriched utterance embedding $\mathbf{\hat{X}}^i_j \in \mathbb{R}^{m \times d}$.

\subsection{Hierarchical Self-Attention}
\begin{table*}[!t]
\small
\centering
\begin{tabular}{cccccc}
 \hline
 \textbf{Dataset} & \textbf{Domain} & \textbf{\#Conv. (Train/Val/Test)}& \textbf{\#Utter. (Train/Val/Test)}& \textbf{\#Classes}& \textbf{Evaluation}\\
 \hline
 EC &Tweet&30160/2755/5509&90480/8265/16527&4&Micro-F1\\
 DailyDialog &Daily Communication&11118/1000/1000&87170/8069/7740&7&Micro-F1\\
 MELD &TV Show Scripts&1038/114/280&9989/1109/2610&7&Weighted-F1\\
 EmoryNLP &TV Show Scripts&659/89/79&7551/954/984&7&Weighted-F1\\
 IEMOCAP &Emotional Dialogues&100/20/31&4810/1000/1523&6&Weighted-F1\\
 \hline
\end{tabular}
\caption{Dataset descriptions.}
\label{table: dataset}
\end{table*}
We propose a hierarchical self-attention mechanism to exploit the structural representation of conversations and learn a vector representation for the contextual utterances $X^i_{j-1}, ..., X^i_{j-M}$. Specifically, the hierarchical self-attention follows two steps: 1) each utterance representation is computed using an utterance-level self-attention layer, and 2) a context representation is computed from $M$ learned utterance representations using a context-level self-attention layer. 

At step 1, for each utterance $X^i_n$, $n$=$j-1$, ..., $j-M$, its representation $\mathbf{\hat{X}}^{'i}_n \in \mathbb{R}^{m \times d}$ is learned as follows:
\begin{equation}
\label{eqn: MultiHead Output}
\mathbf{\hat{X}}^{'i}_n = \textit{FF}(\textit{L}^{'}(\textit{MH}(\textit{L}(\mathbf{\hat{X}}^i_n), \textit{L}(\mathbf{\hat{X}}^i_n), \textit{L}(\mathbf{\hat{X}}^i_n)))),
\end{equation}
where $\textit{L}(\mathbf{\hat{X}}^i_n) \in \mathbb{R}^{m \times h\times d_s}$ is linearly transformed from $\mathbf{\hat{X}}^i_n$ to form $h$ heads ($d_s = d/h$), $\textit{L}^{'}$ linearly transforms from $h$ heads back to 1 head, and 
\begin{align}
    \textit{MH}(Q,K,V) = \textit{softmax}(\frac{QK^T}{\sqrt{d_s}})V,\\
    \textit{FF}(x) = \textit{max}(0, xW_1 + b_1)W_2 + b_2,
\end{align}
where $Q$, $K$, and $V$ denote sets of queries, keys and values, respectively, $W_1 \in \mathbb{R}^{d\times p}, b_1 \in \mathbb{R}^p, W_2 \in \mathbb{R}^{p\times d}$ and $b_2 \in \mathbb{R}^d$ denote model parameters, and $p$ denotes the hidden size of the point-wise feedforward layer (FF) \cite{vaswani2017attention}. The multi-head self-attention layer (MH) enables our model to jointly attend to information from different representation subspaces \cite{vaswani2017attention}. The scaling factor $\frac{1}{\sqrt{d_s}}$ is added to ensure the dot product of two vectors do not get overly large. Similar to \cite{vaswani2017attention}, both MH and FF layers are followed by residual connection and layer normalization, which are omitted in Equation \ref{eqn: MultiHead Output} for brevity.

At step 2, to effectively combine all utterance representations in the context, the context-level self-attention layer is proposed to hierarchically learn the context-level representation $\mathbf{C}^i \in \mathbb{R}^{M \times m \times d}$ as follows:
\begin{equation}
\mathbf{C}^i = \textit{FF}(\textit{L}^{'}(\textit{MH}(\textit{L}(\mathbf{\hat{X}}^i), \textit{L}(\mathbf{\hat{X}}^i), \textit{L}(\mathbf{\hat{X}}^i)))),
\end{equation}
where $\mathbf{\hat{X}}^i$ denotes $[\mathbf{\hat{X}}^{'i}_{j-M}; ...; \mathbf{\hat{X}}^{'i}_{j-1}]$, which is the concatenation of all learned utterance representations in the context.

\subsection{Context-Response Cross-Attention}
Finally, a context-aware concept-enriched response representation $\mathbf{R}^i \in \mathbb{R}^{m \times d}$ for conversation $X^i$ is learned by cross-attention \cite{bahdanau2014neural}, which selectively attends to the concept-enriched context representation as follows:
\begin{equation}
\mathbf{R}^i = \textit{FF}(\textit{L}^{'}(\textit{MH}(\textit{L}(\mathbf{\hat{X}}^{'i}_j), \textit{L}(\mathbf{C}^i), \textit{L}(\mathbf{C}^i)))),
\end{equation}
where the response utterance representation $\mathbf{\hat{X}}^{'i}_j \in \mathbb{R}^{m \times d}$ is obtained via the MH layer:
\begin{equation}
\mathbf{\hat{X}}^{'i}_j = \textit{L}^{'}(\textit{MH}(\textit{L}(\mathbf{\hat{X}}^i_j), \textit{L}(\mathbf{\hat{X}}^i_j), \textit{L}(\mathbf{\hat{X}}^i_j))),
\end{equation}

The resulted representation $\textbf{R}^i \in \mathbb{R}^{m \times d}$ is then fed into a max-pooling layer to learn discriminative features among the positions in the response and derive the final representation $\textbf{O} \in \mathbb{R}^d$:
\begin{equation}
\label{eqn: max pool}
\textbf{O} = \textit{max\_pool}(\textbf{R}^i).
\end{equation}
The output probability $p$ is then computed as
\begin{equation}
\label{eqn: output}
p = \textit{softmax}(\textbf{O}W_3 + b_3),
\end{equation}
where $W_3 \in \mathbb{R}^{d \times q}$ and $b_3 \in \mathbb{R}^q$ denote model parameters, and $q$ denotes the number of classes. The entire KET model is optimized in an end-to-end manner as defined in Equation \ref{eqn: objective}. Our model is available at here\footnote{https://github.com/zhongpeixiang/KET}.

\begin{table*}[!t]
\small
\centering
\begin{tabular}{cccccc}
 \hline
 \textbf{Model} & \textbf{EC} & \textbf{DailyDialog} & \textbf{MELD} & \textbf{EmoryNLP} & \textbf{IEMOCAP}\\
 \hline
 cLSTM & 0.6913 & 0.4990 & 0.4972 & 0.2601 & 0.3484\\
 CNN \cite{kim2014convolutional} & 0.7056 & 0.4934 & 0.5586 & 0.3259 & 0.5218\\
 CNN+cLSTM \cite{poria2017context} & 0.7262 & 0.5024 & 0.5687 & 0.3289 & 0.5587\\
 BERT\_BASE \cite{devlin2018bert} & 0.6946 & 0.5312 & 0.5621 & 0.3315 & 0.6119\\
 DialogueRNN \cite{majumder2018dialoguernn} & 0.7405 & 0.5065 & 0.5627 & 0.3170 & \textbf{0.6121}\\
 \hline
 KET\_SingleSelfAttn (ours) & 0.7285 & 0.5192 & 0.5624 & 0.3251 & 0.5810\\
 KET\_StdAttn (ours) & \textbf{0.7413} & 0.5254 & 0.5682 & 0.3353 & 0.5861\\
 KET (ours) & 0.7348 & \textbf{0.5337} & \textbf{0.5818} & \textbf{0.3439} & 0.5956\\
 \hline
\end{tabular}
\caption{Performance comparisons on the five test sets. Best values are highlighted in bold.}
\label{table: performance comparisons}
\end{table*}

\begin{table}[!t]
\small
\centering
\begin{tabular}{cccccc}
 \hline
 \textbf{Dataset} & \textbf{M}& \textbf{m} & \textbf{d}& \textbf{p}& \textbf{h}\\
 \hline
 EC & 2& 30 & 200 & 100 & 4\\
 DailyDialog & 6& 30 & 300 & 400 & 4\\
 MELD & 6& 30 & 200 & 100 & 4\\
 EmoryNLP & 6& 30 & 100 & 200 & 4\\
 IEMOCAP & 6& 30 & 300 & 400 & 4\\
 \hline
\end{tabular}
\caption{Hyper-parameter settings for KET. $M$: context length. $m$: number of tokens per utterance. $d$: word embedding size. $p$: hidden size in FF layer. $h$: number of heads.}
\label{table: hyper-parameters}
\end{table}
\section{Experimental Settings}
\label{experiments}
In this section we present the datasets, evaluation metrics, baselines, our model variants, and other experimental settings.
\subsection{Datasets and Evaluations}
\label{sec: dataset and evaluations}
We evaluate our model on the following five emotion detection datasets of various sizes and domains. The statistics are reported in Table \ref{table: dataset}. 

\noindent\textbf{EC} \cite{chatterjee2019understanding}: Three-turn Tweets. The emotion labels include happiness, sadness, anger and other.

\noindent\textbf{DailyDialog} \cite{li2017dailydialog}: Human written daily communications. The emotion labels include neutral and Ekman's six basic emotions \cite{ekman1992argument}, namely happiness, surprise, sadness, anger, disgust and fear.

\noindent\textbf{MELD} \cite{poria2018meld}: TV show scripts collected from \textit{Friends}. The emotion labels are the same as the ones used in DailyDialog.

\noindent\textbf{EmoryNLP} \cite{zahiri2018emotion}: TV show scripts collected from \textit{Friends} as well. However, its size and annotations are different from MELD. The emotion labels include neutral, sad, mad, scared, powerful, peaceful, and joyful.

\noindent\textbf{IEMOCAP} \cite{busso2008iemocap}: Emotional dialogues. The emotion labels include neutral, happiness, sadness, anger, frustrated, and excited.


In terms of the evaluation metric, for EC and DailyDialog, we follow \cite{chatterjee2019understanding} to use the micro-averaged F1 excluding the majority class (neutral), due to their extremely unbalanced labels (the percentage of the majority class in the test set is over 80\%). For the rest relatively balanced datasets, we follow \cite{majumder2018dialoguernn} to use the weighted macro-F1. 

\subsection{Baselines and Model Variants}
\label{sec: baselines}
For a comprehensive performance evaluation, we compare our model with the following baselines:

\noindent\textbf{cLSTM}: A contextual LSTM model. An utterance-level bidirectional LSTM is used to encode each utterance. A context-level unidirectional LSTM is used to encode the context.

\noindent\textbf{CNN} \cite{kim2014convolutional}: A single-layer CNN with strong empirical performance. This model is trained on the utterance-level without context.

\noindent\textbf{CNN+cLSTM} \cite{poria2017context}: An CNN is used to extract utterance features. An cLSTM is then applied to learn context representations.


\noindent\textbf{BERT\_BASE} \cite{devlin2018bert}: Base version of the state-of-the-art model for sentiment classification. We treat each utterance with its context as a single document. We limit the document length to the last 100 tokens to allow larger batch size. We do not experiment with the large version of BERT due to memory constraint of our GPU.

\noindent\textbf{DialogueRNN} \cite{majumder2018dialoguernn}: The state-of-the-art model for emotion detection in textual conversations. It models both context and speakers information. The CNN features used in DialogueRNN are extracted from the carefully tuned CNN model. For datasets without speaker information, i.e., EC and DailyDialog, we use two speakers only. For MELD and EmoryNLP, which have 260 and 255 speakers, respectively, we additionally experimented with clipping the number of speakers to the most frequent ones (6 main speakers + an universal speaker representing all other speakers) and reported the best results.

\noindent\textbf{KET\_SingleSelfAttn}: We replace the hierarchical self-attention by a single self-attention layer to learn context representations. Contextual utterances are concatenated together prior to the single self-attention layer.

\noindent\textbf{KET\_StdAttn}: We replace the dynamic context-aware affective graph attention by the standard graph attention \cite{veličković2018graph}.

\subsection{Other Experimental Settings}
\label{sec: experimental settings}
We preprocessed all datasets by lower-casing and tokenization using Spacy\footnote{https://spacy.io/}. We keep all tokens in the vocabulary\footnote{We keep tokens with minimum frequency of 2 for DailyDialog due to its large vocabulary size}. We use the released code for BERT\_BASE and DialogueRNN. For each dataset, all models are fine-tuned based on their performance on the validation set. 

For our model in all datasets, we use Adam optimization \cite{kingma2014adam} with a batch size of 64 and learning rate of 0.0001 throughout the training process. We use GloVe embedding \cite{pennington2014glove} for initialization in the word and concept embedding layers\footnote{We use GloVe embeddings from Magnitude Medium: https://github.com/plasticityai/magnitude}. 
For the class weights in cross-entropy loss for each dataset, we set them as the ratio of the class distribution in the validation set to the class distribution in the training set. Thus, we can alleviate the problem of unbalanced dataset.
The detailed hyper-parameter settings for KET are presented in Table \ref{table: hyper-parameters}. 
\section{Result Analysis}
\label{result analysis}
In this section we present model evaluation results, model analysis, and error analysis.

\begin{figure*}[!t]
\centering
\includegraphics[scale=0.6]{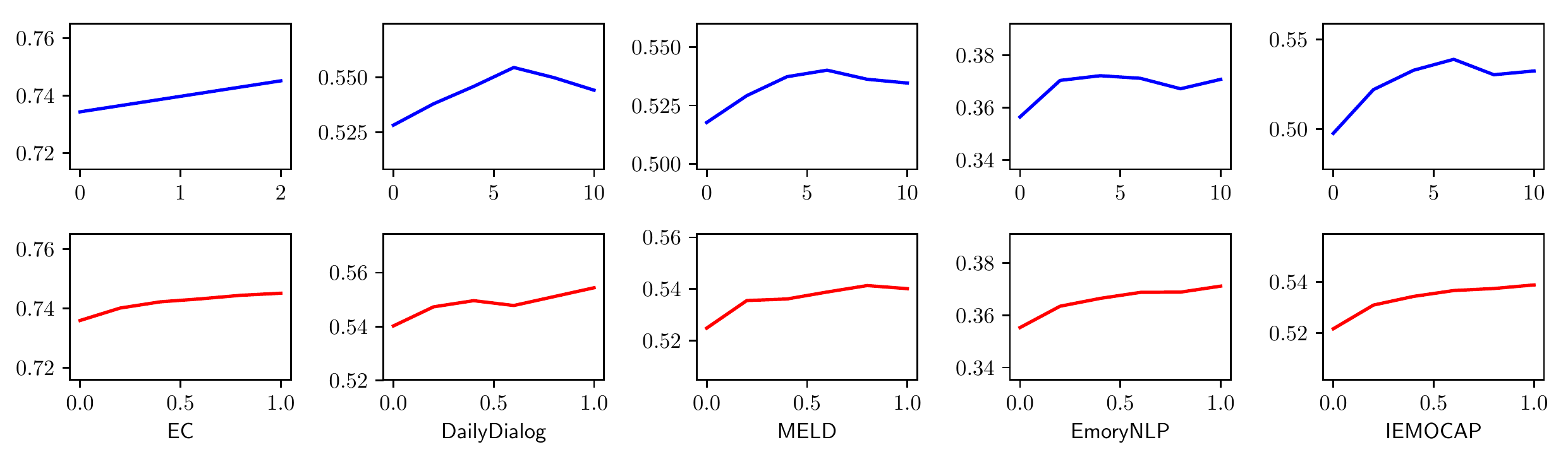}
\caption{Validation performance by KET. Top: different context length ($M$). Bottom: different sizes of random fractions of ConceptNet.}
\label{fig: KBs}
\end{figure*}
\subsection{Comparison with Baselines}
\label{sec: comparison with baselines}
We compare the performance of KET against that of the baseline models on the five afore-introduced datasets. The results are reported in Table \ref{table: performance comparisons}. Note that our results for CNN, CNN+cLSTM and DialogueRNN on EC, MELD and IEMOCAP are slightly different from the reported results in \cite{majumder2018dialoguernn, poria2019emotion}.

cLSTM performs reasonably well on short conversations (i.e., EC and DailyDialog), but the worst on long conversations (i.e., MELD, EmoryNLP and IEMOCAP). One major reason is that learning long dependencies using gated RNNs may not be effective enough because the gradients are expected to propagate back through inevitably a huge number of utterances and tokens in sequence, which easily leads to the vanishing gradient problem \cite{bengio1994learning}. In contrast, when the utterance-level LSTM in cLSTM is replaced by features extracted by CNN, i.e., the CNN+cLSTM, the model performs significantly better than cLSTM on long conversations, which further validates that modelling long conversations using only RNN models may not be sufficient.
BERT\_BASE achieves very competitive performance on all datasets except EC due to its strong representational power via bi-directional context modelling using the Transformer. Note that BERT\_BASE has considerably more parameters than other baselines and our model (110M for BERT\_BASE versus 4M for our model), which can be a disadvantage when deployed to devices with limited computing power and memory.
The state-of-the-art DialogueRNN model performs the best overall among all baselines. In particular, DialogueRNN performs better than our model on IEMOCAP, which may be attributed to its detailed speaker information for modelling the emotion dynamics in each speaker as the conversation flows.

It is encouraging to see that our KET model outperforms the baselines on most of the datasets tested. This finding indicates that our model is robust across datasets with varying training sizes, context lengths and domains. 
Our KET variants KET\_SingleSelfAttn and KET\_StdAttn perform comparably with the best baselines on all datasets except IEMOCAP. However, both variants perform noticeably worse than KET on all datasets except EC, validating the importance of our proposed hierarchical self-attention and dynamic context-aware affective graph attention mechanism. One observation worth mentioning is that these two variants perform on a par with the KET model on EC. Possible explanations are that 1) hierarchical self-attention may not be critical for modelling short conversations in EC, and 2) the informal linguistic styles of Tweets in EC, e.g., misspelled words and slangs, hinder the context representation learning in our graph attention mechanism.

\subsection{Model Analysis}
\label{sec: model analysis}

We analyze the impact of different settings on the validation performance of KET. All results in this section are averaged over 5 random seeds.

\noindent\textbf{Analysis of context length:} 
We vary the context length $M$ and plot model performance in Figure \ref{fig: KBs} (top portion). Note that EC has only a maximum number of 2 contextual utterances. It is clear that incorporating context into KET improves performance on all datasets. However, adding more context is contributing diminishing performance gain or even making negative impact in some datasets. This phenomenon has been observed in a prior study \cite{su2018time}. One possible explanation is that incorporating long contextual information may introduce additional noises, e.g., polysemes expressing different meanings in different utterances of the same context. More thorough investigation of this diminishing return phenomenon is a worthwhile direction in the future.

\noindent\textbf{Analysis of the size of ConceptNet:} 
We vary the size of ConceptNet by randomly keeping only a fraction of the concepts in ConceptNet when training and evaluating our model. The results are illustrated in Figure \ref{fig: KBs} (bottom portion). Adding more concepts consistently improves model performance before reaching a plateau, validating the importance of commonsense knowledge in detecting emotions. We may expect the performance of our KET model to improve with the growing size of ConceptNet in the future.
\begin{table}[!t]
\small
\centering
\begin{tabular}{ccccc}
 \hline
 \textbf{Dataset} & \textbf{0} & \textbf{0.3} & \textbf{0.7} & \textbf{1}\\
 \hline
 EC & 0.7345 & 0.7397 & \textbf{0.7426} & 0.7363\\
 DailyDialog & 0.5365 & 0.5432 & \textbf{0.5451} & 0.5383\\
 MELD & 0.5321 & \textbf{0.5395} & 0.5366 & 0.5306\\
 EmoryNLP & 0.3528 & \textbf{0.3624} & 0.3571 & 0.3488\\
 IEMOCAP & 0.5344 & \textbf{0.5367} & 0.5314 & 0.5251\\
 \hline
\end{tabular}
\caption{Analysis of the relatedness-affectiveness tradeoff on the validation sets. Each column corresponds to a fixed $\lambda_k$ for all concepts (see Equation \ref{eqn: tradeoff}).}
\label{table: relatedness-affectiveness tradeoff}
\end{table}
\begin{table}[!t]
\small
\centering
\begin{tabular}{cccc}
 \hline
 \textbf{Dataset} & KET & -context & -knowledge\\
 \hline
 EC & \textbf{0.7451} & 0.7343 & 0.7359\\
 DailyDialog & \textbf{0.5544} & 0.5282 & 0.5402\\
 MELD & \textbf{0.5401} & 0.5177 & 0.5248\\
 EmoryNLP & \textbf{0.3712} & 0.3564 & 0.3553\\
 IEMOCAP & \textbf{0.5389} & 0.4976 & 0.5217\\
 \hline
\end{tabular}
\caption{Ablation study for KET on the validation sets.}
\label{table: ablation study}
\end{table}

\noindent\textbf{Analysis of the relatedness-affectiveness tradeoff:}
\label{sec: tradeoff}
We experiment with different values of $\lambda_k \in [0,1]$ (see Equation \ref{eqn: tradeoff}) for all $k$ and report the results in Table \ref{table: relatedness-affectiveness tradeoff}. It is clear that $\lambda_k$ makes a noticeable impact on the model performance. Discarding relatedness or affectiveness completely will cause significant performance drop on all datasets, with one exception of IEMOCAP. One possible reason is that conversations in IEMOCAP are emotional dialogues, therefore, the affectiveness factor in our proposed graph attention mechanism can provide more discriminative power.

\noindent\textbf{Ablation Study:}
We conduct ablation study to investigate the contribution of context and knowledge as reported in Table \ref{table: ablation study}. It is clear that both context and knowledge are essential to the strong performance of KET on all datasets. Note that removing context has a greater impact on long conversations than short conversations, which is expected because more contextual information is lost in long conversations.

\subsection{Error Analysis}
\label{sec: error analysis}
Despite the strong performance of our model, it still fails to detect certain emotions on certain datasets. We rank the F1 score of each emotion per dataset and investigate the emotions with the worst scores. We found that disgust and fear are generally difficult to detect and differentiate. For example, the F1 score of fear emotion in MELD is as low as 0.0667. One possible cause is that these two emotions are intrinsically similar. The VAD values of both emotions have low valence, high arousal and low dominance \cite{mehrabian1996pleasure}. Another cause is the small amount of data available for these two emotions. How to differentiate intrinsically similar emotions and how to effectively detect emotions using limited data are two challenging directions in this field.

\section{Conclusion}
\label{conclusions}
We present a knowledge-enriched transformer to detect emotions in textual conversations. 
Our model learns structured conversation representations via hierarchical self-attention and dynamically refers to external, context-aware, and emotion-related knowledge entities from knowledge bases. Experimental analysis demonstrates that both contextual information and commonsense knowledge are beneficial to model performance. The tradeoff between relatedness and affectiveness plays an important role as well. In addition, our model outperforms the state-of-the-art models on most of the tested datasets of varying sizes and domains. 

Given that there are similar emotion lexicons to NRC\_VAD in other languages and ConceptNet is a multilingual knowledge base, our model can be easily adapted to other languages. In addition, given that NRC\_VAD is the only emotion-specific component, our model can be adapted as a generic model for conversation analysis.

\section*{Acknowledgments}
The authors would like to thank the anonymous reviewers for their valuable comments. This research is supported, in part, by the National Research Foundation, Prime Minister’s Office, Singapore under its AI Singapore Programme (Award Number: AISG-GC-2019-003) and under its NRF Investigatorship Programme (NRFI Award No. NRF-NRFI05-2019-0002). This research is also supported, in part, by the Alibaba-NTU Singapore Joint Research Institute, Nanyang Technological University, Singapore.
\bibliographystyle{acl_natbib}
\bibliography{emnlp-ijcnlp-2019}
\end{document}